\documentclass{article}
\usepackage[preprint]{neurips_2026}
\usepackage[T1]{fontenc}
\usepackage[utf8]{inputenc}
\usepackage{microtype}
\usepackage{amsmath,amssymb}
\usepackage{graphicx}
\usepackage{booktabs}
\usepackage{array}
\usepackage{hyperref}
\usepackage{url}
\hypersetup{colorlinks=true,linkcolor=black,citecolor=black,urlcolor=black,
pdftitle={TurboMamba: Faster Inference, Faster Training, and Better Prediction with Lexical Memory},
pdfauthor={Hanxin Zhang},pdfsubject={Blockwise state-space language modeling},
pdfkeywords={language models, Mamba, efficient inference, lexical memory}}
\newcommand{\model}{TurboMamba}

\title{TurboMamba: Faster Inference, Faster Training,\protect\\and Better Prediction with Lexical Memory}
\author{Hanxin Zhang\\
Department of Neurology, University of Pittsburgh\\
zhangh20@upmc.edu}

\begin{document}
\maketitle

\begin{abstract}
We introduce \model{}, a language-model architecture for faster inference, faster training, and better token prediction. A global Mamba-3 model runs once per completed token block. A small Transformer uses this global context and earlier tokens within the block to predict the next token. Three lightweight additions improve local predictions: direct access to the previous block's last token, a small correction to next-token scores, and a learned mixture with recent-token frequencies. We compare twelve small-scale implementations on public FineWeb-Edu, with comparable parameter counts and shared training-budget caps. The baselines include Transformer, Mamba-3, Jamba, and Zamba. The Jamba and Zamba versions use Mamba-3 in place of their original Mamba modules. \model{} achieves the lowest mean test prediction loss (negative log-likelihood, NLL) among all twelve models. Its mean NLL is $9.4\%$ lower than that of Mamba-3, $7.3\%$ lower than that of the Jamba version, and $18.3\%$ lower than that of the Zamba version. End-to-end training throughput is $2.60\times$ that of Mamba-3, $2.81\times$ that of the Jamba version, and $4.82\times$ that of the Zamba version. We measure end-to-end batch latency for a batch of 32 sequences, each with a 256-token prompt and 128 generated tokens. On this workload, \model{} achieves a $3.12\times$ speedup over Mamba-3, a $3.53\times$ speedup over the Jamba version, and a $6.21\times$ speedup over the Zamba version.
\end{abstract}

\section{Introduction}
Every generated text unit, or token, has two costs: carrying forward the context and producing an accurate prediction. Transformers retain detailed context through attention. State-space models such as Mamba carry it in a compact recurrent state \citep{gu2023mamba,lahoti2026mamba3}. Hierarchical models reduce work further by processing groups of tokens with a global model and predicting the tokens with a smaller local model \citep{yu2023megabyte,ho2024block,egli2025multiscale}.

\model{} follows a simple design principle: \textbf{spend global computation on blocks, and spend a small amount of extra computation on token detail}. A global Mamba-3 stack processes completed token blocks. A lightweight local Transformer predicts each token using the global context and the tokens already available inside the block. A boundary connection, a residual vocabulary correction, and a recent-count mixture enrich this local prediction path. Model width, block size, stack depths, and memory length define a configuration of the architecture; the experiment below evaluates one small-scale configuration.

We call our basic comparison model the \emph{Plain block model}. It uses the same block-level Mamba-3 and local Transformer stacks, but omits the previous-block token connection, residual vocabulary correction, and recent-token mixture. Each block's first token is predicted from its global prefix; later tokens also use earlier tokens in that block.

Our contributions are threefold. First, we specify and jointly train this lexical enhancement package within a blockwise Mamba-3 architecture. Second, we implement incremental generation with block-level state updates and token-level lexical updates. Third, we compare twelve implementations spanning attention, state-space models, layer hybrids, hierarchies, and lexical memory under a common public-data and budget protocol. We measure prediction quality, end-to-end training throughput, generated-text diagnostics, and end-to-end batch latency across five batch sizes.

\section{Related Work}
\paragraph{Hierarchical language models.}
MEGABYTE uses global and local Transformers over fixed byte patches, including a cross-patch attention option \citep{yu2023megabyte}. Block Transformer applies global-to-local modeling to tokens and supplies coarse context to a local decoder \citep{ho2024block}. Multiscale Byte Language Model (MBLM) generalizes the hierarchy to multiple stages and explicitly studies a global Mamba with a local Transformer \citep{egli2025multiscale}. It is the closest prior architectural skeleton to the Plain block model. H-Net learns dynamic byte chunks and combines fine-resolution processing with a coarser main model; it also describes cached autoregressive execution \citep{hwang2025hnet}. BLT allocates global computation to entropy-based byte patches \citep{pagnoni2024blt}. Block-State Transformer (BST) instead combines a token-resolution state-space model with block-local attention \citep{fathi2023bst}.

\paragraph{Layer hybrids.}
Jamba combines token-level Mamba and attention layers, with top-2 expert routing in selected feedforward layers \citep{lieber2024jamba}. This routing selects two feedforward subnetworks, called experts, for each token. Zamba reuses a shared attention and feedforward branch throughout a Mamba backbone \citep{glorioso2024zamba}. These designs mix mechanisms at token resolution. \model{} instead reduces the global update rate and enriches the small local prediction path.

\begin{table}[t]
\centering
\small
\setlength{\tabcolsep}{3pt}
\renewcommand{\arraystretch}{1.06}
\caption{Mechanism comparison. The rows describe architectural designs; the experiments use matched small-scale implementations. TurboMamba combines blockwise global computation with three additions to its local prediction path.}
\label{tab:mechanisms}
\begin{tabular}{>{\raggedright\arraybackslash}p{.17\linewidth}>{\raggedright\arraybackslash}p{.24\linewidth}>{\raggedright\arraybackslash}p{.25\linewidth}>{\raggedright\arraybackslash}p{.24\linewidth}}
\toprule
Architecture & Global computation & Local prediction & Added detail or routing\\
\midrule
\textbf{TurboMamba} & Mamba-3 per fixed block & Small Transformer with global prefix & \textbf{Boundary connection, residual head, count mixture}\\
Jamba & Token-level Mamba and attention & Token-level hybrid layers & Top-2 experts in selected layers\\
Zamba & Token-level Mamba & Repeated shared attention branch & Original-embedding concatenation; shared attention/feedforward\\
BST & Token-level state-space model & Block-local self/cross-attention & State-space context in local attention\\
MBLM & Fixed multiscale hierarchy & Local decoder with global context vectors & Separate stage embeddings\\
H-Net & Learned dynamic chunks & Fine encoder and decoder around main model & Learned boundaries; smoothed chunk expansion\\
BT & Attention per fixed block & Local Transformer with global prefixes & Coarse-to-fine token modeling\\
Cache Transformer & Token-level attention & Common Transformer & Hidden-similarity lexical cache\\
Pointer Sentinel & Token-level attention & Common Transformer & Learned pointer/vocabulary mixture\\
Plain block model & Mamba-3 per fixed block & Small Transformer with global prefix & Three local enhancements removed\\
Mamba-3 & Recurrent update per token & Token-level state-space stack & Compact recurrent context\\
Transformer & Full causal attention & Token-level attention stack & Detailed attention history\\
\bottomrule
\end{tabular}
\end{table}

\paragraph{Lexical memory.}
Lexical memory stores information about recent tokens. Count caches reuse them according to frequency. Continuous Cache weights stored tokens by hidden-state similarity, with a count distribution as a special case \citep{grave2016cache}. Pointer Sentinel learns attention over recent positions and a sentinel that controls the vocabulary mixture \citep{merity2016pointer}. \model{} uses recent-token counts with a bounded learned mixing weight, jointly trained at every prediction position. Its contribution is the particular integration of boundary information, output correction, and lexical memory into the inexpensive local path of a blockwise Mamba-3 model.

\section{The TurboMamba Architecture}
\begin{figure}[t]
\centering
\includegraphics[width=\linewidth]{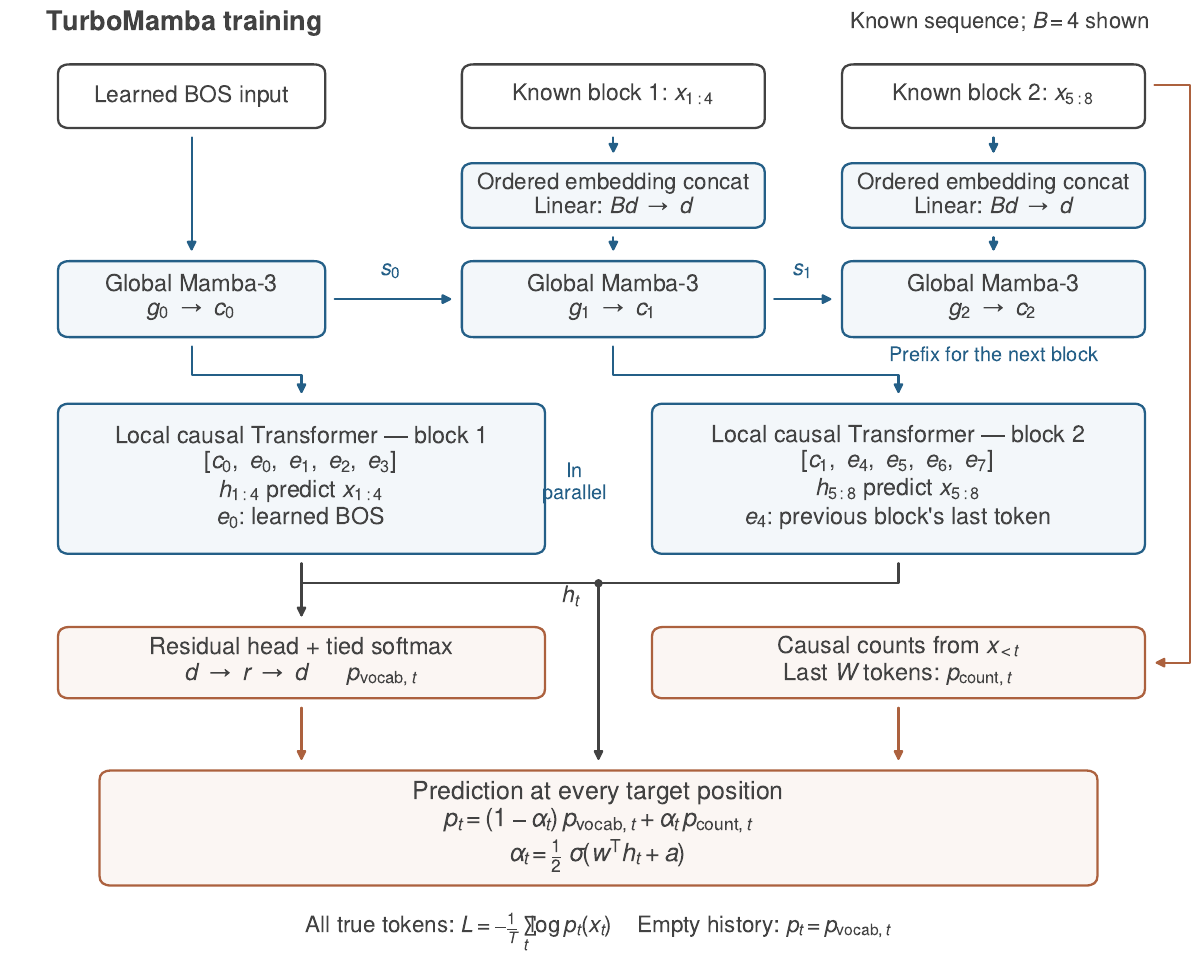}
\caption{Training uses a known token sequence. The global scan is causal over $B$-token blocks, and its outputs are shifted by one block so each target block receives only earlier global context. Once these prefixes are available, the local causal decoders for different blocks can run in parallel. Every target uses earlier token inputs and counts of earlier tokens.}
\label{fig:training}
\end{figure}

Figure~\ref{fig:training} shows the causal global context and parallel local block evaluations for a known training sequence.

\paragraph{Blockwise global context.}
Let $x_t$ be a token and $e_t=E[x_t]\in\mathbb{R}^{d}$ its embedding, a learned vector. For completed block $j$ of length $B$, we concatenate its embeddings in order and project them to the global width:
\begin{equation}
 b_j=W_b[e_{(j-1)B+1};\ldots;e_{jB}],\qquad
 (g_j,s_j)=G(b_j,s_{j-1}).
 \label{eq:global}
\end{equation}
Here $G$ is the global Mamba-3 stack, $s_j$ denotes its persistent state, and $g_j$ is its normalized output. A learned projection produces one global prefix $c_j=W_p g_j$, a context vector prepended to the next local block. A learned beginning-of-sequence input supplies the first prefix. During training, the prefix is shifted by one block so each target receives context from completed earlier blocks.

\paragraph{Boundary-aware local prediction.}
For block $j$, let $a_j=(j-1)B+1$ be its first token index. Causal prediction uses only earlier tokens. At position $k\in\{1,\ldots,B\}$, the local Transformer $L$ receives
\begin{equation}
 h_{j,k}=L\big([c_{j-1};e_{a_j-1};e_{a_j:a_j+k-2}]\big)_{\rm last}.
 \label{eq:local}
\end{equation}
The slice is empty for $k=1$. The first token therefore has direct access to the previous block's last raw-token embedding $e_{a_j-1}$. Later tokens also attend to it through the local keys and values. For the first block, a learned beginning-of-sequence vector takes its place. $L$ includes the local stack's final RMS normalization. The local attention cache resets at each new block.

\paragraph{Residual vocabulary correction.}
Before the tied output projection, a small bottleneck adds a learned correction to the local representation:
\begin{equation}
 \widetilde h_t=h_t+W_2\,\mathrm{GELU}\!\left(W_1\,\mathrm{RMSNorm}(h_t)\right),
 \qquad p_{\rm vocab,t}=\mathrm{softmax}(E\widetilde h_t).
 \label{eq:head}
\end{equation}
The bottleneck has dimensions $d\rightarrow r\rightarrow d$. Input embeddings and the vocabulary projection share weights.

\paragraph{Recent-token mixture.}
The memory contains only already observed tokens. With memory length $W$ and $H_t=\{\max(1,t-W),\ldots,t-1\}$, the final distribution is
\begin{equation}
 \begin{aligned}
 p_{\rm count,t}(v)&=\frac{1}{|H_t|}\sum_{i\in H_t}\mathbf{1}[x_i=v],\\
 \alpha_t&=\tfrac12\sigma(w^\top h_t+a),\\
 p_t(v)&=(1-\alpha_t)p_{\rm vocab,t}(v)+\alpha_t p_{\rm count,t}(v).
 \end{aligned}
 \label{eq:mixture}
\end{equation}
When history is empty, $p_t=p_{\rm vocab,t}$. The count gate $\alpha_t$ is a learned mixing weight. It uses the local representation before the residual correction and assigns at most half the probability mass to token counts. All components learn together through $\mathcal{L}=-T^{-1}\sum_{t=1}^T\log p_t(x_t)$.

\section{Training and Incremental Inference}
\begin{figure}[t]
\centering
\includegraphics[width=\linewidth]{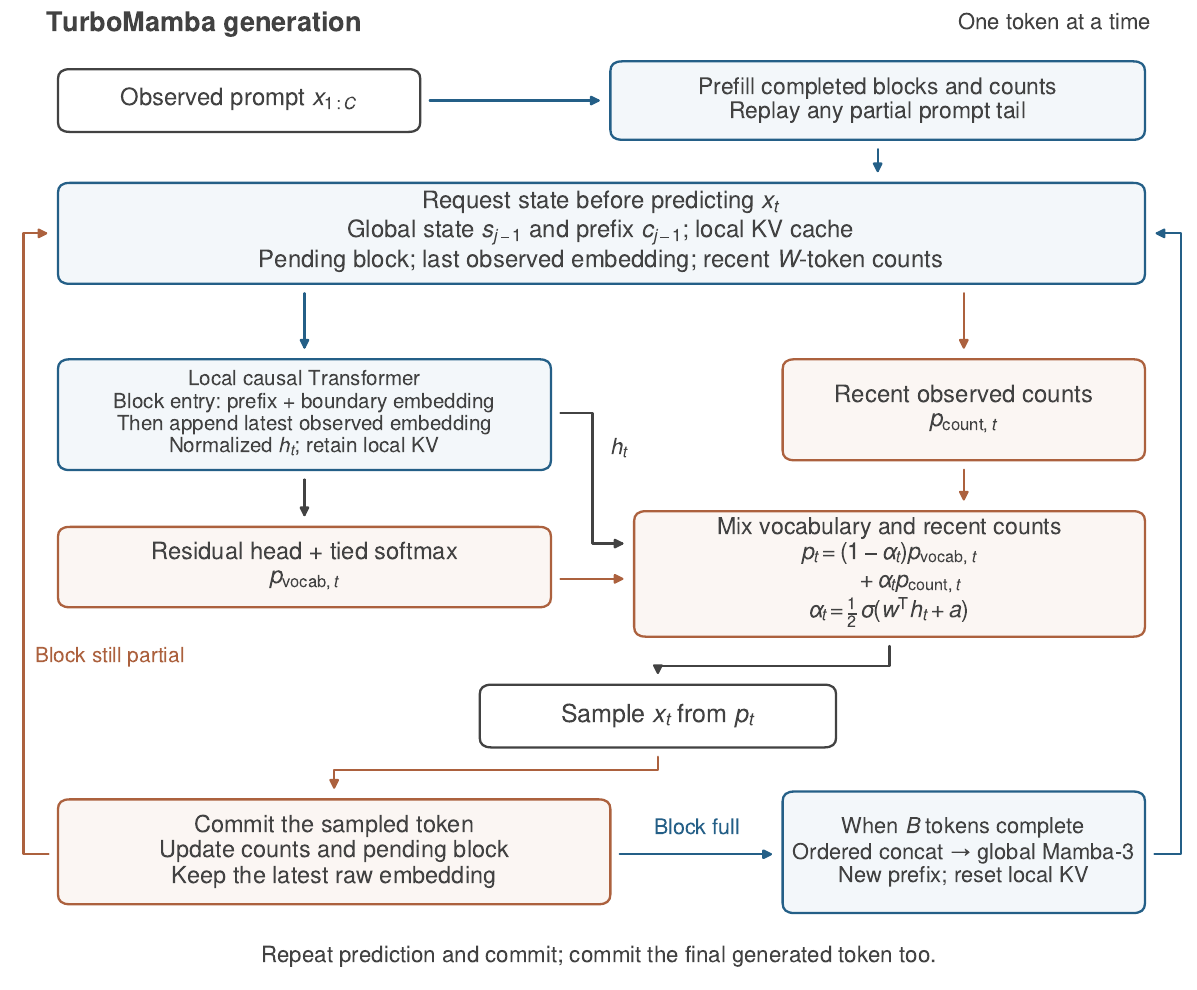}
\caption{Generation predicts and commits one token at a time. The local prediction path runs at every token; each commit updates recent counts and the partial block. The global Mamba-3 state advances only after a complete $B$-token block, then the local cache resets. Sampling remains sequential. The final generated token is committed, including any completed-block global update.}
\label{fig:generation}
\end{figure}

Known training sequences allow the global scan to process the shortened block sequence and the local decoders to process blocks in parallel, as in Figure~\ref{fig:training}. Every token contributes to the mixed likelihood. Figure~\ref{fig:generation} shows the sequential state updates used at inference.

During generation, each sequence has a global state, a local key/value (KV) cache storing attention keys and values from earlier inputs, a partial block, and a recent-token count buffer. Prefill processes the prompt to initialize these states. Each token prediction evaluates the local path and mixture; a commit then consumes the sampled token and updates recent counts and the partial block. Completing the $B$th token advances the global state and starts a fresh local cache. The last generated token is committed as part of the request.

This schedule reduces global recurrent updates to one per $B$ tokens. The persistent global state, local cache, partial block, and lexical memory have bounded size with respect to generated history. We check full-sequence and cached log probabilities for every model and seed, along actual sampled histories and after the final token commit.

\section{Experimental Protocol}
\paragraph{Small-scale architecture study.}
We evaluate one reference configuration with approximately three million parameters. It uses $B=4$, $d=128$, $r=64$, and $W=128$, with six global Mamba-3 single-input/single-output (SISO) layers and two local Transformer layers. Local attention has four query heads, two key/value heads, and rotary positions. Feedforward width is 509, vocabulary size is 4,096, and the exact parameter count is 3,000,753. This configuration supplies the initial empirical study of the \model{} architecture.

\paragraph{Public text and prediction task.}
We use a fixed subset of FineWeb-Edu's public 10BT sample \citep{penedo2024fineweb}. Documents are partitioned by website host; duplicate URLs, normalized texts, and shared long fragments are screened before tokenization. A 4,096-entry byte-level byte-pair encoding (BPE) tokenizer is fitted on training text. The training bank contains 10,898 within-document windows with 256 scored targets each, or 2,789,888 target positions. The development (DEV, validation) and test sets contain 881 and 1,024 fixed continuation windows. Test NLL scores 128 true continuation tokens after a 256-token context, equally weighting documents within each seed. It uses natural logarithms, with units of nats per target token; lower NLL means better next-token prediction on this task. This expanded study reuses the earlier study's data split; all current model and adapter choices use development data.

\paragraph{Models and budget.}
We compare twelve implementations: a causal Transformer, token-level Mamba-3 SISO, Block Transformer (BT), BST with S4D (a diagonal state-space model), a Continuous Cache Transformer, a Pointer Sentinel Transformer, a plain global-Mamba/local-Transformer control, Jamba (Mamba-3 adaptation) and Zamba (Mamba-3 adaptation) hybrids, MBLM, H-Net, and \model{}. Total parameter counts are matched within 0.56\%. All share the tokenizer, data, training batch 16, and 256-target sequences. The Cache model uses the corresponding freshly trained Transformer final checkpoint and is charged its complete backbone training cost. A fresh DEV-only grid search selects cache interpolation on the selected current Transformer development pilot.

The Jamba and Zamba implementations retain their layer-hybrid mechanisms and use the common Mamba-3 core. MBLM uses two stages and adds global context vectors to local inputs; H-Net uses learned boundaries, a coarse main model, and causal smoothing when expanding chunks back to token resolution. Both use the shared BPE vocabulary. Appendix~\ref{sec:details} specifies the scaled configurations. All eleven trainable architectures receive current development selection and formal training from initialization under the same training protocol. Quality, generated text, and end-to-end batch latency are freshly measured for all twelve under the same inference procedure. The shared environment and precision are specified in Appendix~\ref{sec:environment}.

Every trained architecture receives two learning-rate trials (0.001 and 0.003), each capped at 30 seconds, followed by three formal seeds (9201, 9202, and 9203), each capped at 180 seconds: 22 development pilots and 33 formal training runs in total. A random seed determines model initialization and training data order. The cap includes initialization, data staging, development scoring, and checkpoint work. AdamW uses $(0.9,0.95)$ momentum coefficients, weight decay 0.1 on all parameters, gradient clipping at 1, and a ten-update warmup. Formal runs start from fresh model and optimizer states; selection uses development data, and final checkpoints supply test results. Faster models process more targets within this shared wall-clock budget. The maximum/minimum total training and model-selection cost is 1.066; Table~\ref{tab:training_cost} reports actual DEV, formal, and total costs.

\paragraph{End-to-end training throughput.}
This metric divides scored training target tokens, including revisits, by the complete formal training job's wall-clock time from launch until completion is detected. The clock includes startup, model and data staging, all training steps, development evaluation, checkpointing, and exit. Shared dataset preparation and tokenizer fitting occur beforehand; learning-rate selection trials and Cache selection are separately accounted in Table~\ref{tab:training_cost}. Cache throughput includes its full Transformer backbone training. Auxiliary \emph{Training-step throughput} covers batch staging, forward and backward computation, gradient clipping, optimizer updates, and synchronization. For each metric, we report the mean and sample standard deviation (SD) of the three per-seed token/time ratios.

\paragraph{End-to-end batch latency.}
This metric is the wall-clock duration of one fixed batch of $n$ independent sequences, each with a $C$-token prompt and $G$ generated tokens. It is measured in milliseconds per batch, without dividing by $n$ or $G$. We use $n\in\{1,4,8,16,32\}$ and two workloads: $(C,G)=(256,128)$ and $(128,64)$. The clock includes input copying and transfer of the prompt and pre-generated random numbers used for sampling, input checks, prefill, sequential temperature-$0.8$/top-$p$-$0.9$ sampling, all state updates and the final-token commit, output transfer and checks, and completion synchronization. Random-number creation occurs before the clock; weight loading and cold setup are measured separately. For each model, seed, workload, and batch size, four full-batch warmups prepare the runtime and are excluded from the reported statistics, followed by twelve measured batches in a balanced twelve-model order. Each batch starts with fresh sequence states. We take the median of the twelve durations within each seed and report the mean of the three seed medians.

\section{Results}
\begin{table}[t]
\centering
\small
\caption{Prediction quality, Training-step throughput, and End-to-end training throughput. Values are means $\pm$ sample SD across three seeds. Both throughput metrics use thousands of scored target tokens/s; End-to-end training throughput covers the complete formal job. Bold numbers identify the best metric-column means.}
\label{tab:quality}
\begin{tabular}{lrrrr}
\toprule
Model & Parameters (M) & Test NLL $\downarrow$ & \shortstack{Training-step\\throughput $\uparrow$} & \shortstack{End-to-end\\training\\throughput $\uparrow$}\\
\midrule
Transformer & 3.000 & $5.290 \pm 0.127$ & $18.29 \pm 2.86$ & $17.50 \pm 2.79$ \\
Mamba-3 SISO & 3.001 & $5.455 \pm 0.056$ & $8.01 \pm 0.08$ & $6.83 \pm 0.07$ \\
BT & 3.000 & $5.201 \pm 0.013$ & $\mathbf{23.56} \pm 0.23$ & $\mathbf{22.67} \pm 0.22$ \\
BST (S4D) & 3.000 & $5.638 \pm 0.010$ & $10.06 \pm 0.14$ & $9.45 \pm 0.14$ \\
Cache Transformer & 3.000 & $5.001 \pm 0.116$ & $18.29 \pm 2.86$ & $17.50 \pm 2.79$ \\
Pointer Sentinel & 3.016 & $5.185 \pm 0.098$ & $16.14 \pm 0.37$ & $15.41 \pm 0.38$ \\
Plain block model & 2.999 & $5.366 \pm 0.021$ & $20.27 \pm 0.18$ & $19.06 \pm 0.19$ \\
\textbf{TurboMamba} & 3.001 & $\mathbf{4.941} \pm 0.028$ & $18.92 \pm 0.46$ & $17.77 \pm 0.45$ \\
\shortstack[l]{Jamba\\{\footnotesize(Mamba-3 adaptation)}} & 3.000 & $5.328 \pm 0.069$ & $7.46 \pm 0.03$ & $6.33 \pm 0.04$ \\
\shortstack[l]{Zamba\\{\footnotesize(Mamba-3 adaptation)}} & 3.000 & $6.044 \pm 0.063$ & $4.83 \pm 0.18$ & $3.69 \pm 0.16$ \\
MBLM & 3.002 & $6.110 \pm 0.114$ & $21.00 \pm 3.07$ & $19.77 \pm 3.00$ \\
H-Net & 3.001 & $5.349 \pm 0.036$ & $17.16 \pm 0.60$ & $16.10 \pm 0.59$ \\
\bottomrule
\end{tabular}
\end{table}

\begin{figure}[t]
\centering
\includegraphics[width=\linewidth]{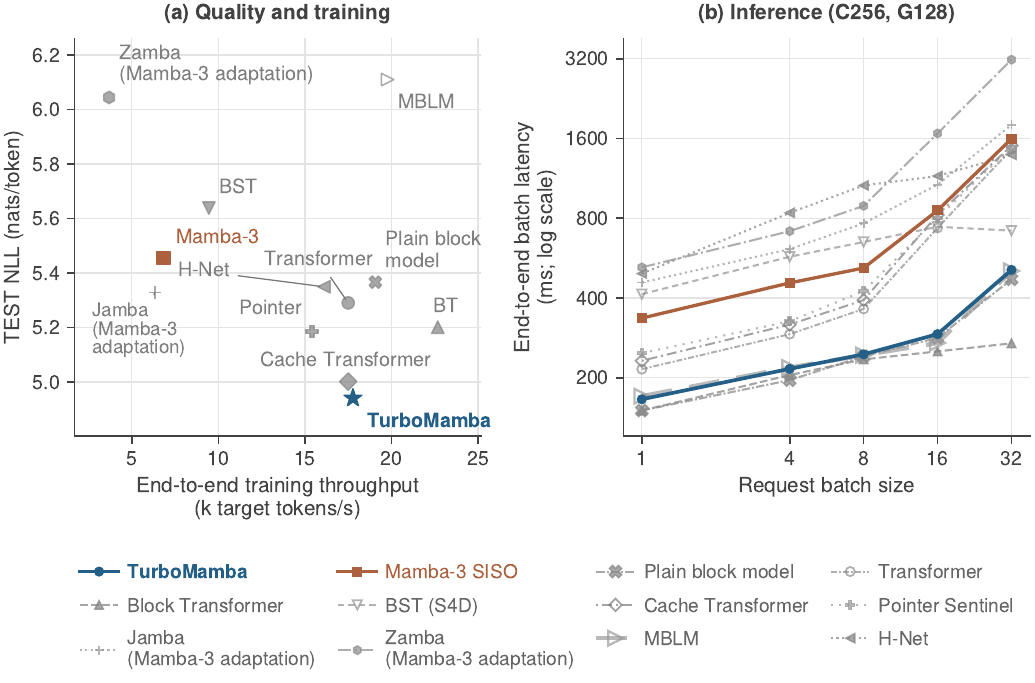}
\caption{Test NLL and End-to-end training throughput (left); End-to-end batch latency for 256 prompt tokens and 128 generated tokens per sequence (right). Both panels include all twelve implementations. Each point or curve summarizes three formal seeds. Blue and rust highlight TurboMamba and Mamba-3. The other models, including the Plain block model, use gray lines and markers, with different line styles and marker shapes.}
\label{fig:results}
\end{figure}

\paragraph{Quality and End-to-end training throughput.}
\model{} obtains the lowest mean test NLL in Table~\ref{tab:quality}. Its NLL is 4.941 $\pm$ 0.028, compared with 5.455 $\pm$ 0.056 for Mamba-3 SISO, 5.201 $\pm$ 0.013 for BT, and 5.001 $\pm$ 0.116 for the Cache Transformer. End-to-end training throughput is 17.77k targets/s, $2.60\times$ that of Mamba-3. BT and MBLM reach 22.67k and 19.77k targets/s, respectively. Figure~\ref{fig:results} shows the twelve-model quality--throughput comparison.

\begin{table}[t]
\centering
\small
\caption{End-to-end batch latency (ms per batch, lower is better). Each of the $n$ independent sequences has a 256-token prompt and 128 generated tokens. Values are means of three per-seed medians. Bold numbers identify the lowest mean in each batch column.}
\label{tab:latency}
\begin{tabular}{lrrrrr}
\toprule
Model & $n=1$ & $n=4$ & $n=8$ & $n=16$ & $n=32$\\
\midrule
Transformer & $215.6$ & $292.3$ & $363.8$ & $735.4$ & $1457.1$ \\
Mamba-3 SISO & $336.0$ & $456.2$ & $519.3$ & $856.3$ & $1595.5$ \\
BT & $\mathbf{150.5}$ & $204.8$ & $\mathbf{234.8}$ & $\mathbf{251.6}$ & $\mathbf{269.9}$ \\
BST (S4D) & $413.8$ & $572.6$ & $650.9$ & $743.1$ & $718.5$ \\
Cache Transformer & $232.0$ & $317.7$ & $393.0$ & $824.1$ & $1503.5$ \\
Pointer Sentinel & $247.9$ & $326.9$ & $424.8$ & $803.5$ & $1486.0$ \\
Plain block model & $150.6$ & $\mathbf{196.0}$ & $241.2$ & $282.9$ & $470.8$ \\
\textbf{TurboMamba} & $166.0$ & $216.2$ & $245.3$ & $292.4$ & $510.6$ \\
\shortstack[l]{Jamba\\{\footnotesize(Mamba-3 adaptation)}} & $457.9$ & $611.6$ & $765.4$ & $1071.3$ & $1801.8$ \\
\shortstack[l]{Zamba\\{\footnotesize(Mamba-3 adaptation)}} & $522.3$ & $715.1$ & $890.7$ & $1671.5$ & $3173.4$ \\
MBLM & $170.5$ & $218.6$ & $243.6$ & $270.8$ & $506.2$ \\
H-Net & $494.6$ & $838.5$ & $1063.7$ & $1154.8$ & $1388.2$ \\
\bottomrule
\end{tabular}
\end{table}

\paragraph{End-to-end batch latency across batch sizes.}
For the 256-prompt/128-generation workload, \model{} has End-to-end batch latency of 166.0 ms at $n=1$ and 510.6 ms at $n=32$. The Mamba-3/TurboMamba End-to-end batch latency ratio is 2.93 at $n=16$ and 3.12 at $n=32$. Across both workloads and all five batch sizes, the comparator/TurboMamba ratios span Mamba-3 SISO: 1.99--3.12; BT: 0.46--0.96; MBLM: 0.93--1.03; H-Net: 2.33--4.34. A ratio above one favors TurboMamba. Tables~\ref{tab:latency} and \ref{tab:secondary_latency} report all twelve implementations.

\section{What the Complete Design Adds}
The Plain block model defined in the Introduction provides the basic blockwise backbone. Comparing it with \model{} measures the combined contribution of the previous-block token connection, residual vocabulary correction, and recent-token mixture. Both variants use six global Mamba-3 layers and two local Transformer layers in this experiment. The Plain block model has feedforward hidden width 514, compared with 509 for \model{}, to keep their total parameter counts matched.

The package lowers NLL from 5.366 for the Plain block model to 4.941, a reduction of 0.426 nats per token. The three seed improvements are 0.423, 0.416, 0.437; the exploratory paired seed/document 95\% interval for the reduction is $[0.410,0.442]$. This comparison evaluates the three additions together under matched total capacity. Appendix~\ref{sec:details} reports paired contrasts against all controls and separate generation diagnostics.

The architectural contribution is the placement and joint operation of the three additions. The boundary connection brings a precise recent token into the new block. The bottleneck enriches the vocabulary readout. The count mixture supplies exact recent frequencies through a learned gate. Together, they let the model retain a compact global context while giving the local decoder several inexpensive sources of detail.

\section{Discussion and Conclusion}
\model{} follows a simple design principle: perform global computation less frequently and strengthen local predictions. In this small-scale public-data study, it achieves a mean NLL of 4.941 and End-to-end training throughput of 17.77k target tokens/s. End-to-end batch latency is 510.6 ms for $n=32$ sequences, each with a 256-token prompt and 128 generated tokens. The twelve-model comparisons show how this design balances prediction quality, End-to-end training throughput, and End-to-end batch latency across the measured workloads and batch sizes.

The next stages of the series will examine larger models, longer training, additional datasets, and the contribution of each component. The compact implementation provides a starting point for these extensions. Code is prepared for release at \url{https://github.com/hanxinzhang/turbomamba}.

\begingroup
\small
\setlength{\bibsep}{3pt}
\bibliographystyle{plainnat}
\bibliography{references}

@article{gu2023mamba,
  title={Mamba: Linear-Time Sequence Modeling with Selective State Spaces},
  author={Gu, Albert and Dao, Tri},
  year={2023}, journal={arXiv preprint arXiv:2312.00752},
  url={https://arxiv.org/abs/2312.00752}
}

@article{lahoti2026mamba3,
  title={Mamba-3: Improved Sequence Modeling using State Space Principles},
  author={Lahoti, Aakash and Li, Kevin Y. and Chen, Berlin and Wang, Caitlin and Bick, Aviv and Kolter, J. Zico and Dao, Tri and Gu, Albert},
  year={2026}, journal={arXiv preprint arXiv:2603.15569},
  url={https://arxiv.org/abs/2603.15569}
}

@article{yu2023megabyte,
  title={{MEGABYTE}: Predicting Million-byte Sequences with Multiscale Transformers},
  author={Yu, Lili and Simig, D{\'a}niel and Flaherty, Colin and Aghajanyan, Armen and Zettlemoyer, Luke and Lewis, Mike},
  year={2023}, journal={arXiv preprint arXiv:2305.07185},
  url={https://arxiv.org/abs/2305.07185}
}

@article{ho2024block,
  title={Block Transformer: Global-to-Local Language Modeling for Fast Inference},
  author={Ho, Namgyu and Bae, Sangmin and Kim, Taehyeon and Jo, Hyunjik and Kim, Yireun and Schuster, Tal and Fisch, Adam and Thorne, James and Yun, Se-Young},
  year={2024}, journal={arXiv preprint arXiv:2406.02657},
  url={https://arxiv.org/abs/2406.02657}
}

@article{egli2025multiscale,
  title={Multiscale Byte Language Models -- A Hierarchical Architecture for Causal Million-Length Sequence Modeling},
  author={Egli, Eric and Manica, Matteo and Born, Jannis},
  year={2025}, journal={arXiv preprint arXiv:2502.14553},
  url={https://arxiv.org/abs/2502.14553}
}

@article{hwang2025hnet,
  title={Dynamic Chunking for End-to-End Hierarchical Sequence Modeling},
  author={Hwang, Sukjun and Wang, Brandon and Gu, Albert},
  year={2025}, journal={arXiv preprint arXiv:2507.07955},
  url={https://arxiv.org/abs/2507.07955}
}

@article{pagnoni2024blt,
  title={Byte Latent Transformer: Patches Scale Better Than Tokens},
  author={Pagnoni, Artidoro and Pasunuru, Ram and Rodriguez, Pedro and Nguyen, John and Muller, Benjamin and Li, Margaret and Zhou, Chunting and Yu, Lili and Weston, Jason and Zettlemoyer, Luke and Ghosh, Gargi and Lewis, Mike and Holtzman, Ari and Iyer, Srinivasan},
  year={2024}, journal={arXiv preprint arXiv:2412.09871},
  url={https://arxiv.org/abs/2412.09871}
}

@article{fathi2023bst,
  title={Block-State Transformers},
  author={Fathi, Mahan and Pilault, Jonathan and Firat, Orhan and Pal, Christopher and Bacon, Pierre-Luc and Goroshin, Ross},
  year={2023}, journal={arXiv preprint arXiv:2306.09539},
  url={https://arxiv.org/abs/2306.09539}
}

@article{grave2016cache,
  title={Improving Neural Language Models with a Continuous Cache},
  author={Grave, Edouard and Joulin, Armand and Usunier, Nicolas},
  year={2016}, journal={arXiv preprint arXiv:1612.04426},
  url={https://arxiv.org/abs/1612.04426}
}

@article{merity2016pointer,
  title={Pointer Sentinel Mixture Models},
  author={Merity, Stephen and Xiong, Caiming and Bradbury, James and Socher, Richard},
  year={2016}, journal={arXiv preprint arXiv:1609.07843},
  url={https://arxiv.org/abs/1609.07843}
}

@article{penedo2024fineweb,
  title={The {FineWeb} Datasets: Decanting the Web for the Finest Text Data at Scale},
  author={Penedo, Guilherme and Kydl{\'i}{\v c}ek, Hynek and Ben Allal, Loubna and Lozhkov, Anton and Mitchell, Margaret and Raffel, Colin and von Werra, Leandro and Wolf, Thomas},
  year={2024}, journal={arXiv preprint arXiv:2406.17557},
  url={https://arxiv.org/abs/2406.17557}
}

@article{lieber2024jamba,
  title={Jamba: A Hybrid Transformer-Mamba Language Model},
  author={Lieber, Opher and Lenz, Barak and Bata, Hofit and Cohen, Gal and Osin, Jhonathan and Dalmedigos, Itay and Safahi, Erez and Meirom, Shaked and Belinkov, Yonatan and Shalev-Shwartz, Shai and Abend, Omri and Alon, Raz and Asida, Tomer and Bergman, Amir and Glozman, Roman and Gokhman, Michael and Manevich, Avashalom and Ratner, Nir and Rozen, Noam and Shwartz, Erez and Zusman, Mor and Shoham, Yoav},
  year={2024},
  journal={arXiv preprint arXiv:2403.19887},
  url={https://arxiv.org/abs/2403.19887},
  doi={10.48550/arXiv.2403.19887}
}

@article{glorioso2024zamba,
  title={Zamba: A Compact 7B SSM Hybrid Model},
  author={Glorioso, Paolo and Anthony, Quentin and Tokpanov, Yury and Whittington, James and Pilault, Jonathan and Ibrahim, Adam and Millidge, Beren},
  year={2024},
  journal={arXiv preprint arXiv:2405.16712},
  url={https://arxiv.org/abs/2405.16712},
  doi={10.48550/arXiv.2405.16712}
}
\endgroup

\appendix
\section{Additional Experimental Details}
\label{sec:details}
\subsection{Environment and implementation}
\label{sec:environment}
All neural computation uses an Apple M4 with 24\,GiB memory, PyTorch 2.13.0, and Apple MPS in FP32 with CPU fallback disabled. All comparators use this shared environment and precision. Training uses Torch autograd. Evaluation and generation use Torch RMS normalization, activations, and the common Mamba-3 SISO recurrent step with gradient recording disabled. Inference projections own their contiguous storage; these copies are included in End-to-end batch latency. The pretrained GPT-2 generated-text evaluator also runs on MPS.

Jamba evaluates all four experts at each expert site in this MPS implementation and combines the selected top two outputs. Total and routed active capacity are recorded separately. Synchronization in the End-to-end training throughput and End-to-end batch latency clocks includes completion of the corresponding GPU work.

\subsection{Scaled configurations and additional results}
\paragraph{Scaled hybrid configurations.}
Jamba (Mamba-3 adaptation) uses eight token-level layers: one attention layer and seven Mamba-3 layers. Four sites contain four experts with top-2 routing and a loss that encourages balanced expert use. Total capacity is 2,999,992 parameters; routed active capacity is 2,364,088, counting shared layers and selected experts. Zamba (Mamba-3 adaptation) uses twelve Mamba-3 layers and one shared attention/feedforward branch injected at two sites, with separate site projections and attention histories; its total capacity is 2,999,904.

MBLM uses six global Mamba-3 layers and two local Transformer layers, separate stage embeddings, four-token patches, global context vectors added to local inputs, and the original prediction shifts across flattened stages; it has 3,001,520 parameters. H-Net uses one token-level Mamba-3 encoder layer, six chunk-level Transformer layers, and one token-level Mamba-3 decoder layer. It retains learned chunk boundaries, a straight-through estimator for routing confidence, causal exponential-moving-average smoothing when expanding chunks, a residual connection at token resolution, and an auxiliary loss targeting four input tokens per chunk on average. It has 3,001,232 parameters. All four implementations use the common optimizer and learning-rate grid.

\begin{table}[ht]
\centering
\small
\caption{Recorded training and model-selection costs (s). DEV is the two-pilot total; formal is the three-seed training total used for End-to-end training throughput. Cache total includes 4.250 s of adapter selection in addition to its full Transformer backbone training. Entries are rounded to one decimal.}
\label{tab:training_cost}
\begin{tabular}{lrrr}
\toprule
Model & DEV pilots (s) & Formal training (s) & Total (s)\\
\midrule
Transformer & $52.3$ & $528.8$ & $581.1$ \\
Mamba-3 SISO & $50.7$ & $513.3$ & $564.0$ \\
BT & $52.4$ & $527.6$ & $580.0$ \\
BST (S4D) & $52.1$ & $525.0$ & $577.1$ \\
Cache Transformer & $52.3$ & $528.8$ & $585.3$ \\
Pointer Sentinel & $52.8$ & $528.3$ & $581.1$ \\
Plain block model & $52.8$ & $524.7$ & $577.5$ \\
\textbf{TurboMamba} & $52.9$ & $524.3$ & $577.3$ \\
\shortstack[l]{Jamba\\{\footnotesize(Mamba-3 adaptation)}} & $50.2$ & $509.7$ & $559.9$ \\
\shortstack[l]{Zamba\\{\footnotesize(Mamba-3 adaptation)}} & $49.1$ & $500.1$ & $549.2$ \\
MBLM & $51.7$ & $525.4$ & $577.1$ \\
H-Net & $51.6$ & $522.4$ & $574.0$ \\
\bottomrule
\end{tabular}
\end{table}

\begin{table}[ht]
\centering
\small
\caption{End-to-end batch latency (ms per batch) for $n$ independent sequences, each with a 128-token prompt and 64 generated tokens. Same aggregation and clock boundaries as Table~\ref{tab:latency}. Bold numbers identify the lowest mean in each batch column.}
\label{tab:secondary_latency}
\begin{tabular}{lrrrrr}
\toprule
Model & $n=1$ & $n=4$ & $n=8$ & $n=16$ & $n=32$\\
\midrule
Transformer & $111.0$ & $127.5$ & $164.3$ & $225.5$ & $408.6$ \\
Mamba-3 SISO & $173.9$ & $200.6$ & $262.1$ & $380.3$ & $774.7$ \\
BT & $\mathbf{77.2}$ & $\mathbf{89.4}$ & $113.1$ & $\mathbf{141.4}$ & $\mathbf{119.5}$ \\
BST (S4D) & $211.6$ & $249.2$ & $313.0$ & $392.3$ & $327.5$ \\
Cache Transformer & $117.5$ & $136.4$ & $174.6$ & $236.7$ & $430.0$ \\
Pointer Sentinel & $122.8$ & $141.5$ & $183.5$ & $244.9$ & $435.5$ \\
Plain block model & $78.6$ & $90.6$ & $\mathbf{113.1}$ & $149.1$ & $249.3$ \\
\textbf{TurboMamba} & $87.4$ & $100.7$ & $127.4$ & $159.1$ & $262.4$ \\
\shortstack[l]{Jamba\\{\footnotesize(Mamba-3 adaptation)}} & $236.4$ & $267.0$ & $362.0$ & $474.9$ & $866.8$ \\
\shortstack[l]{Zamba\\{\footnotesize(Mamba-3 adaptation)}} & $266.4$ & $303.7$ & $418.7$ & $699.6$ & $1446.3$ \\
MBLM & $90.0$ & $102.6$ & $130.5$ & $161.6$ & $265.5$ \\
H-Net & $222.0$ & $338.9$ & $450.7$ & $580.8$ & $611.1$ \\
\bottomrule
\end{tabular}
\end{table}

\begin{table}[ht]
\centering
\small
\caption{Paired test NLL differences (nats per target token), \model{} minus comparator. Exploratory 95\% intervals use paired seed/document resampling of the three-seed experiment.}
\label{tab:quality_contrasts}
\begin{tabular}{lrr}
\toprule
Comparator & Mean difference & 95\% interval\\
\midrule
Transformer & $-0.349$ & $[-0.422, -0.228]$ \\
Mamba-3 SISO & $-0.515$ & $[-0.544, -0.483]$ \\
BT & $-0.261$ & $[-0.281, -0.237]$ \\
BST (S4D) & $-0.698$ & $[-0.734, -0.659]$ \\
Cache Transformer & $-0.061$ & $[-0.130, +0.043]$ \\
Pointer Sentinel & $-0.244$ & $[-0.358, -0.130]$ \\
Plain block model & $-0.426$ & $[-0.442, -0.410]$ \\
\shortstack[l]{Jamba\\{\footnotesize(Mamba-3 adaptation)}} & $-0.388$ & $[-0.471, -0.285]$ \\
\shortstack[l]{Zamba\\{\footnotesize(Mamba-3 adaptation)}} & $-1.103$ & $[-1.188, -1.043]$ \\
MBLM & $\mathbf{-1.169}$ & $[-1.254, -1.089]$ \\
H-Net & $-0.409$ & $[-0.463, -0.342]$ \\
\bottomrule
\end{tabular}
\end{table}

\paragraph{Generated-text diagnostics.}
We generate 128 tokens for each of 16 fixed prompts per model and seed (576 samples total). A pretrained GPT-2 scores the generated continuation's negative log-probability, normalized by GPT-2 tokens or UTF-8 bytes. Table~\ref{tab:generation} reports these external-likelihood diagnostics and word four-gram repetition: the fraction of four-consecutive-word sequences that repeat an earlier occurrence. True-continuation NLL is reported in Table~\ref{tab:quality}. \model{} has GPT-2 NLL/byte of 1.802, compared with 1.971 for Mamba-3, and no repeated word four-grams in its 48 samples.

\begin{table}[ht]
\centering
\small
\caption{Generated-text diagnostics, three-seed means. GPT-2 NLL uses nats per GPT-2 token or UTF-8 byte. It and word four-gram repetition are lower-is-better proxies; true-continuation prediction is in Table~\ref{tab:quality}.}
\label{tab:generation}
\begin{tabular}{lrrr}
\toprule
Model & GPT-2 NLL/token & GPT-2 NLL/byte & Word 4-gram repeat (\%)\\
\midrule
Transformer & $7.254$ & $1.927$ & $\mathbf{0.000}$ \\
Mamba-3 SISO & $6.307$ & $1.971$ & $0.030$ \\
BT & $6.364$ & $1.746$ & $\mathbf{0.000}$ \\
BST (S4D) & $6.366$ & $1.878$ & $\mathbf{0.000}$ \\
Cache Transformer & $6.518$ & $\mathbf{1.731}$ & $0.968$ \\
Pointer Sentinel & $\mathbf{6.049}$ & $1.810$ & $5.044$ \\
Plain block model & $6.466$ & $1.856$ & $0.047$ \\
\textbf{TurboMamba} & $6.398$ & $1.802$ & $\mathbf{0.000}$ \\
\shortstack[l]{Jamba\\{\footnotesize(Mamba-3 adaptation)}} & $6.199$ & $1.974$ & $\mathbf{0.000}$ \\
\shortstack[l]{Zamba\\{\footnotesize(Mamba-3 adaptation)}} & $6.815$ & $2.197$ & $\mathbf{0.000}$ \\
MBLM & $6.902$ & $2.092$ & $\mathbf{0.000}$ \\
H-Net & $6.478$ & $2.006$ & $0.031$ \\
\bottomrule
\end{tabular}
\end{table}

\end{document}